\documentclass[11pt,a4paper]{article}
\usepackage[hyperref]{naaclhlt2019}
\usepackage{times}
\usepackage{latexsym}
\usepackage{caption}

\usepackage{url}

\aclfinalcopy 

\usepackage[utf8]{inputenc} 
\usepackage[T1]{fontenc}    
\usepackage{url}            
\usepackage{booktabs}       
\usepackage{amsfonts}       
\usepackage{nicefrac}       
\usepackage{microtype}      
\usepackage{amsmath}
\usepackage{graphicx}
\usepackage{enumitem}
\usepackage[titletoc,toc,title]{appendix}
\usepackage{nidanfloat}

\DeclareMathOperator{\sgn}{sgn}

\usepackage{algorithm}
\usepackage[noend]{algpseudocode}
\usepackage{array}
\usepackage{booktabs}
\usepackage{multirow}
\usepackage{pbox}
\usepackage{caption}
\usepackage{wrapfig}

\usepackage{xargs}   
\usepackage{xcolor}  

\newcommand*{\V}[1]{\mathbf{#1}}%

\newcommand\Tstrut{\rule{0pt}{2.0ex}}       
\newcommand\Bstrut{\rule[-0.8ex]{0pt}{0pt}} 
\newcommand{\TBstrut}{\Tstrut\Bstrut} 

\newcommand{\ffrac}[2]{\ensuremath{\frac{\displaystyle #1}{\displaystyle #2}}}

\usepackage[colorinlistoftodos,prependcaption,textsize=tiny]{todonotes}
\newcommandx{\unsure}[2][1=]{\todo[linecolor=red,backgroundcolor=red!25,bordercolor=red,#1]{#2}}
\newcommandx{\change}[2][1=]{\todo[linecolor=blue,backgroundcolor=blue!25,bordercolor=blue,#1]{#2}}
\newcommandx{\info}[2][1=]{\todo[linecolor=OliveGreen,backgroundcolor=OliveGreen!25,bordercolor=OliveGreen,#1]{#2}}
\newcommandx{\improvement}[2][1=]{\todo[linecolor=Plum,backgroundcolor=Plum!25,bordercolor=Plum,#1]{#2}}
\newcommandx{\thiswillnotshow}[2][1=]{\todo[disable,#1]{#2}}

\newcommand{\cmmnt}[1]{\ignorespaces}

\usepackage{etoolbox}

\makeatletter
\expandafter\patchcmd\csname\string\algorithmic\endcsname{\itemsep\z@}{\itemsep=0.2ex plus2pt}{}{}
\makeatother

\algrenewcommand\algorithmicindent{0.9em}%

\makeatletter
\newcommand{\algmargin}{\the\ALG@thistlm}
\makeatother
\newlength{\whilewidth}
\algdef{SE}[parWHILE]{parWhile}{EndparWhile}[1]
  {\parbox[t]{\dimexpr\linewidth-\algmargin}{%
     \hangindent\whilewidth\strut\algorithmicwhile\ #1\ \algorithmicdo\strut}}{\algorithmicend\ \algorithmicwhile}%
\algnewcommand{\parState}[1]{\State%
  \parbox[t]{\dimexpr\linewidth-\algmargin}{\strut #1\strut}}

\title{F10-SGD: Fast Training of Elastic-net Linear Models for Text Classification and Named-entity Recognition}

\author{
  Stanislav Peshterliev, Alexander Hsieh, Imre Kiss \\
  Amazon, Alexa Machine Learning, USA \\
  \texttt{\{stanislp,hsiea,ikiss\}@amazon.com} \\
}

\date{}

\hypersetup{draft}

\begin{document}
\maketitle
\begin{abstract}
Voice-assistants text classification and named-entity recognition (NER) models are trained on millions of example utterances. Because of the large datasets, long training time is one of the bottlenecks for releasing improved models. In this work, we develop F10-SGD, a fast optimizer for text classification and NER elastic-net linear models. On internal datasets, F10-SGD provides 4x reduction in training time compared to the OWL-QN optimizer without loss of accuracy or increase in model size. Furthermore, we incorporate biased sampling that prioritizes harder examples towards the end of the training. As a result, in addition to faster training, we were able to obtain statistically significant accuracy improvements for NER.

On public datasets, F10-SGD obtains 22\% faster training time compared to FastText for text classification. And, 4x reduction in training time compared to CRFSuite OWL-QN for NER.
\end{abstract}

\section{Introduction}
\label{sec:intro}

Our commercial voice-assistant natural language understanding (NLU) system uses linear MaxEnt models~\cite{berger1996maximum} for text classification, and linear-chain CRF models~\cite{lafferty2001conditional, sutton2012introduction} for named-entity recognition (NER). There are several desirable properties that makes these models suitable for NLU:
\begin{itemize}[topsep=1pt, leftmargin=15pt, itemsep=-1pt]
\item They achieve high accuracy given the right set of features~\cite{wang2012baselines}.
\item They produce probabilistic output, which allows combining multiple models in a recognition pipeline~\cite{su2018re}.
\item They can be compressed to run on hardware-constrained devices \cite{strimel2018statistical}.
\end{itemize}

To train MaxEnt and CRF models, we minimize an objective function that includes the negative log-likelihood of the training dataset and elastic-net regularization. The elastic-net regularization combines the L1 and L2 penalties on the model weights. The L1 penalty, sum of absolute weights, serves as feature selection mechanism by forcing the unimportant model weights to zero. The L2 penalty, sum of squared weights, helps avoid overfitting by preventing weights that are excessively large.

A popular way to train elastic-net linear models is the Orthant-Wise Limited-memory Quasi-Newton Optimizer (OWL-QN) \cite{andrew2007scalable}. OWL-QN is a variant of the LBFGS \cite{zhu1997algorithm} quasi-Newton optimization method that supports the L1 penalty. OWL-QN produces compact and accurate models, and it can be speed-up using multiple CPU threads. 

In our voice-assistant's NLU, we classify user utterance text into hundreds of classes. We have a hierarchical classification system, where we use MaxEnt to first perform domain classification (DC) and then intent classification (IC). DC predicts the general domain  of an utterance. IC predicts the user intent within a domain. For example, for the utterance ``play music'', DC predict \textit{Music} and IC predict \textit{PlayMusicIntent}, and for the utterance ``how is the weather'', DC predicts \textit{Weather} and IC predicts \textit{GetWeatherIntent}. This NLU architecture where intents are separated by domain allows us to efficiently train the IC models in parallel on their respective domain-specific training data. However, the MaxEnt DC training still has to be done on the entire training dataset. A DC model training with OWL-QN using 28 CPU threads on 50 million utterances takes around 2 hours. 

For NER, we use one CRF model per domain to recognize domain-specific named-entities, such as artist name and song name for the Music domain. For example, in the sentence ``play desert rose by sting'', NER labels ``desert rose'' as a song and ``sting'' as an artist. The CRF models are more complex and slower to train compared to the MaxEnt models. An NER model training with OWL-QN using 28 CPU threads on a large domain with 6 million utterances takes around 2 hours 40 min.

Continuous improvement of the NLU system requires frequent retraining of the DC, IC, and NER models. However, the long training times for DC and NER are a bottleneck. In this work, we focus on machine learning optimization methods to improve training efficiency. We develop a fast optimizer called F10-SGD which on large internal datasets trains MaxEnt and CRF models 4 times faster compared to OWL-QN without loss of accuracy or increase in the model size.

Our contributions are the following:
\begin{itemize}[topsep=1pt, leftmargin=15pt, itemsep=-1pt]
\item We combine Stochastic Gradient Descent (SGD) optimization techniques for parallel training and elastic-net regularization into a fast and accurate trainer.
\item We improve the accuracy of the NER model using biased sampling that prioritizes harder examples towards the end of the training.
\item We perform evaluations on internal and public datasets. On internal datasets, F10-SGD is 4x faster with half the CPU threads compared to OWL-QN. On external datasets, we speed-up training by 22\% and 3.4-3.7x against FastText and CRFSuite respectively.
\end{itemize}

\textbf{Why not GPU?} An important constrain for our use case is to develop a fast optimizer for CPU instead of GPU. Our model building platform trains many models per week, and allocating GPU for each training is expensive.

\section{Related Work}

\label{sec:related}

\citet{bottou2010large,bottou2012stochastic} advocated for using SGD in large scale machine learning optimization. He developed CRFSGD\footnote{\url{http://leon.bottou.org/projects/sgd}} and demonstrated the effectiveness of SGD for training CRF models. \citet{CRFsuite} developed the popular CRFSuite toolkit, which includes an SGD trainer based on CRFSGD, and an optimized implementation of the OWL-QN trainer. A limitation of both CRFSGD and CRFSuite is that they do not use multiple CPU threads to speed up training. Also, their SGD implementations do not support L1 regularization.



 
Scikit-learn~\cite{pedregosa2011scikit} and Vowpal Wabbit~\cite{langford2007vowpal} provide fast SGD implementations for logistic regression models (note that logistic regression is equivalent MaxEnt). Both support L1 regularization using cumulative penalty~\cite{tsuruoka2009stochastic} and truncated gradient \cite{langford2009sparse}. Scikit-learn SGD does not support parallel SGD, and in Vowpal Wabbit parallel SGD is not supported, though it can be simulated via sending raw training examples over sockets to multiple Vowpal Wabbit processes that share the same memory.

FastText~\cite{joulin2016bag} is linear embedding model for text classification. It supports asynchronous multi-threaded SGD training via Hogwild~\cite{recht2011hogwild}, which makes training fast. However, FastText does not support L2 or dropout regularization, leading to suboptimal performance on small datasets. Also, it does not support L1 for feature selection, but it does have a quantization option \cite{joulin2016fasttext} to reduce the model size after training.

\section{Background}

\subsection{Elastic-net Linear Models}

Probabilistic linear models have the general form as shown in Equation~\ref{log_linear_general}, where $\V{x}$ is an input example, $\V{y}$ is a prediction target, $w_i$ is a weight for the feature function $\phi_i$, and $Z$ is a partition function that normalizes the probability. If the prediction target $\V{y}$ is a class as in IC and DC, the model is a MaxEnt. If the prediction target $\V{y}$ is sequence of labels as in NER, the model is a CRF.
\setlength{\abovedisplayskip}{3pt}
\setlength{\belowdisplayskip}{3pt}
\setlength{\abovedisplayshortskip}{0pt}
\setlength{\belowdisplayshortskip}{0pt}
\begin{align}
\label{log_linear_general}
p(\V{x}|\V{y}) &= \frac{1}{Z(\V{x})}\exp \sum_i w_i \phi_i(\V{x},\V{y}) 
\end{align}

Equation~\ref{objective} shows the optimization objective for MaxEnt and CRF. It includes the negative log-likelihood over the training set $D$ with size $N$, and the L1 and L2 penalties.
\setlength{\abovedisplayskip}{3pt}
\setlength{\belowdisplayskip}{3pt}
\setlength{\abovedisplayshortskip}{0pt}
\setlength{\belowdisplayshortskip}{0pt}
\begin{align}
\label{objective}
F(\V{w}) =& \sum_j^{N} L_j(\V{w}) + \lambda_1 \| \V{w} \|_1 + \frac{\lambda_2}{2} \| \V{w} \|_2^2 \\
L_j(\V{w}) &= -\log p(\V{y}_j | \V{x}_j; \V{w})
\end{align}
To optimize $F$, we use the gradient $\nabla F$ in Equation~\ref{gradient}. The function $\sgn$ returns: -1 for negative, 1 positive, or 0 for every weight of $\V{w}$. $\sgn$ is necessary to define the sub-gradient of $F$ for zero weights since the absolute $|\cdot|$ function in the L1 penalty is not differentiable at zero.
\vspace{-2pt}
\begin{align}
\label{gradient}
\nabla F &= \sum_j^N \nabla L_j + \lambda_1 \sgn(\V{w}) +  \lambda_2 \V{w}
\end{align}

The definitions of $\nabla L_j$ and $p(\V{x}|\V{y})$ differ for MaxEnt and CRF, and are not covered in this paper.

\subsection{OWL-QN}

OWL-QN is an optimization method that iteratively adjusts the model parameters towards the optimal value. Algorithm~\ref{owl-qn} provides a high-level simplified description of OWL-QN that helps explain its computational bottlenecks, but does not describe the algorithm in details.
\vspace{-5pt}
\begin{algorithm}[H]

\caption{OWL-QN($F$, $\V{w}^0$)}
\label{owl-qn}

\begin{algorithmic}[1]
\State $V \gets \{\}$, $G \gets \{\}$
\For{e $\gets$ 0,1,..,Epochs or Convergence}
	\State $\V{d} \gets \V{H}^{e} \times \nabla^e F$ using V and G
	\State $\alpha \gets \min_{\alpha \geq 0} F(\V{w}^e - \alpha \V{d})$ 
	\State $\V{w}^{e+1} \gets \V{w}^e - \alpha \V{d}$
	
	\State Update V with $\V{v}^e \gets \V{w}^{e+1} - \V{w}^{e}$
	\State Update G with $\V{g}^e \gets \nabla^{e+1} F - \nabla^e F$
\EndFor

\end{algorithmic}

\end{algorithm}
\vspace{-10pt}
OWL-QN approximates the inverse hessian $\V{H}^e$ using the last $m$ weight differences $V$, and gradient differences $G$. The quality of the approximation $\V{H}^e$ depends on the size $m$, a recommended value for $m$ is between 4 and 7.


In step 3, OWL-QN computes an update direction $\V{d}$ using $\V{H}^e$ and the gradient $\nabla^e F$ over $N$ training examples. Then, in step 4 it performs a line search in the direction $\V{d}$. For a large $N$ even one epoch takes significant time, and OWL-QN typically needs $m$ epochs to find good approximation $\V{H}^e$, and tens or even hundreds of epochs to converge.

\textbf{Parallelization.} We can speed-up OWL-QN by using multiple CPU threads to compute $F$ and $\nabla F$. Each CPU thread $t$ receives a subset of the training dataset $D$ and computes the forward scores $\sum_j^{D_t} L_j$ and the gradients $\sum_j^{D_t} \nabla L_j$. Then, the results are aggregated to compute $F$ and $\nabla F$.

\section{F10-SGD}
\label{sec:approach}

We propose using SGD as a fast alternative to OWL-QN for training MaxEnt and CRF models. Similar to OWL-QN, SGD is an iterative optimization method. It uses the gradient approximation $\nabla F_j$ in Equation~\ref{sgd_gradient}. $\nabla F_j$ is calculated from $\nabla L_j$ for a single random training example. The weights $\V{w}$ are then updated with learning rate $\alpha$, without relying on the inverse hessian. Since $\V{w}$ is updated many times during a single epoch, SGD typically needs fewer epochs to converge compared to OWL-QN. 
\setlength{\abovedisplayskip}{3pt}
\setlength{\belowdisplayskip}{3pt}
\setlength{\abovedisplayshortskip}{0pt}
\setlength{\belowdisplayshortskip}{0pt}
\begin{align}
\label{sgd_gradient}
\nabla F_j &= \nabla L_j + \lambda^\prime_1 \sgn(\V{w}) + \lambda^\prime_2 \V{w} \\
\lambda^\prime_1 &= \frac{\lambda_1}{N} \quad \lambda^\prime_2 = \frac{\lambda_2}{N} \nonumber
\end{align}

In $\nabla F_j$ we divide $\lambda_1$ and $\lambda_2$ by the number of examples $N$ since $\V{w}$ is updated $N$ times per epoch.

F10-SGD in Algorithm~\ref{sgd} is a parallel version of SGD, also known as Hogwild. Each CPU thread $t$ receives a random subset of the training dataset $D$, and updates the weights $\V{w}$. The weight updates are not synchronized, so it is possible for threads to override each other. However, the algorithm has a nearly optimal rate of convergence for models where many training examples have non-overlapping features~\cite{recht2011hogwild}, which is the case for linear MaxEnt and CRF models.

\begin{algorithm}[H]

\caption{F10-SGD($F$, $\V{w}^0$, T)}
\label{sgd}

\begin{algorithmic}[1]
\State atomic k $\gets$ 0
\For{e $\gets$ 0,1,..,Epochs or Convergence}
	\State $S \gets $ RandomShuffle(0..N)	
	\For{t $\gets$ 0,1,..,T}
    \State $S_t \gets S$ from $tN/T$ to $(t + 1)N/T$
	\State Run \Call{SubEpoch}{$S_t$} on new thread
	\EndFor
\EndFor

\Procedure{SubEpoch}{$S_t$}
	\For{$j \in S_t$}
	
	\State $\alpha \gets $ LearningRate(k)
	
	\State UpdateWeights(k, $\alpha$)
	
	\State $k \gets k + 1$
	
	\EndFor
\EndProcedure
\end{algorithmic}

\end{algorithm}



\textbf{Learning Rate.} For SGD to converge, it is critical that the learning rate $\alpha$ is neither too large nor too small. We start with an initial learning rate $\alpha_0$, and decay it linearly during training using $\alpha \gets \alpha_0 (1 - k / N \times \mathit{Epochs})$ \cite{joulin2016bag}. We tune $\alpha_0$ on a held-out development set using grid search.


The learning rate $\alpha$ depends on the iteration counter $k$. To have the same $\alpha$ with multiple threads  across identical runs, we have to update $k$ atomically. Otherwise, $\alpha$ is going to differ, which causes fluctuation of the weights $\V{w}$ and makes it harder to reproduce previous models. The same atomicity consideration applies to the L1 and L2 algorithms below. 



\subsection{Lazy Updates}


For linear MaxEnt and CRF models, $\V{w}$ typically has millions of weights for features corresponding to different words and phrases in the training data. Fortunately, a single example $j$ has only a small set of active features $\phi_i(\V{x}_j, \V{y}_j)$, so we can compute $\nabla L_j$ efficiently using only the weights $i$. However, the computation of $\nabla F_j$ requires applying the L1 and L2 penalties of the full weight vector $\V{w}$ which is slow. Thus, it is essential to perform ``lazy'' L1 and L2 updates only on the active features.

\textbf{Lazy L2.} To make the L2 update lazy, we use the weight rescaling approach in Algorithm~\ref{l2_update_alg} \cite{shalev2011pegasos}. The key observation is that the L2 penalty can be seen as rescaling $\V{w}$ by $(1 - \alpha \lambda^{\prime}_2)$ at every update, Equations~\ref{l2_sparse_update_1}-\ref{l2_sparse_update_2}. Thus, we can represent $\V{w}^k$ as $\V{\hat{w}}^k s_k$ and update the unscaled weights $\V{\hat{w}}^k$ and the scaler $s_k$ independently. At the end of every epoch, we multiply the weights ${\hat{w}}^k_i$ by $s_k$ and reset $s_k \gets 1$. 
\setlength{\abovedisplayskip}{3pt}
\setlength{\belowdisplayskip}{3pt}
\setlength{\abovedisplayshortskip}{0pt}
\setlength{\belowdisplayshortskip}{0pt}
\begin{align}
\label{l2_sparse_update_1}
\V{w}^{k + 1} \gets \V{w}^k - \alpha \nabla^k L_j - \alpha \lambda^{\prime}_2 \V{w}^k \\
\label{l2_sparse_update_2}
\V{w}^{k + 1} \gets \V{w}^k (1 - \alpha \lambda^{\prime}_2) - \alpha \nabla^k L_j
\end{align}


\vspace{-3pt}
\begin{algorithm}[H]

\caption{Sparse L2 Update}
\label{l2_update_alg}

\begin{algorithmic}[1]
\State atomic $s_0 \gets 1$

\Procedure{UpdateWeights}{k, $\alpha$}
	\State $s_{k+1} \gets (1 - \alpha \lambda^\prime_2) s_{k}$ 
	
	\For{$i \in$ features in $j$}
	
	\State $\hat{w}_i^{k + \frac{1}{2}} \gets \hat{w}_i^k - \alpha \nabla^k_i L_j / s_{k + 1}$
	
	\EndFor
	
	\State ApplyL1(k, $\alpha$)
\EndProcedure

\end{algorithmic}

\end{algorithm}

\vspace{-15pt}

\begin{algorithm}[H]
\caption{Sparse L1 Update}
\label{l1_update_alg}

\begin{algorithmic}[1]

\State atomic $u_0 \gets 0$; $\V{q}^0 \gets \V{0}$

\Procedure{ApplyL1}{k, $\alpha$}

\State $u_{k + 1} \gets u_k + \alpha\lambda^\prime_1 / s_{k + 1}$

\For{$i \in$ features in $j$}

\If{$\hat{w}_i^{k + \frac{1}{2}}$ > 0}

    \State $\hat{w}^{k + 1}_i \gets \max (0, \hat{w}_i^{k + \frac{1}{2}} - u_{k + 1} + q_i^k)$
    
\ElsIf{$\hat{w}_i^{k + \frac{1}{2}}$ < 0}

    \State $\hat{w}^{k + 1}_i \gets \min (0, \hat{w}_i^{k + \frac{1}{2}} + u_{k + 1} - q_i^k)$
    
\EndIf

\State $q^{k+1}_i \gets q^k_i + (\hat{w}^{k + 1}_i - \hat{w}_i^{k + \frac{1}{2}}) $

\EndFor

\EndProcedure

\end{algorithmic}

\end{algorithm}
\vspace{-20pt}
\setlength{\abovedisplayskip}{3pt}
\setlength{\belowdisplayskip}{3pt}
\setlength{\abovedisplayshortskip}{0pt}
\setlength{\belowdisplayshortskip}{0pt}
\begin{align}
\label{l1_sparse_update_1}
\V{\hat{w}}^{k + 1} \gets \V{\hat{w}}^{k + \frac{1}{2}} - \alpha \lambda^\prime_1 \sgn(\V{\hat{w}^k}s_{k})
\end{align}
\textbf{Lazy L1}. To make the L1 update lazy, we use the cumulative L1 penalty approach \cite{tsuruoka2009stochastic} in Algorithm~\ref{l1_update_alg}. First, from Equation~\ref{l1_sparse_update_1}, we note that the L1 update rarely makes weights $\V{\hat{w}}$ zero, so it does reduce model size. Thus, we modify the L1 update to clip weights to zero if they cross the zero threshold after applying the L1 penalty. Second, to lazily update a weight $\hat{w}^k_i$, we have to maintain the cumulative L1 penalty $u_k$, the L1 penalty that is already applied to the weight $q^k_i$, and update the weight with the difference, steps 6 and 8 of Algorithm~\ref{l1_update_alg}. At the end of every epoch, we run \texttt{ApplyL1} for every weight $i$ and reset $u_k+1 \gets 0$ and $q^{k+1}_i \gets 0$.

Note that in step 3 we have to divide by the scaler $s_{k+1}$ to boost the L1 penalty in order to account for the L2 regularization.


\subsection{Active Bias Sampling}

Active bias sampling \cite{chang2017active} is a technique that can speed-up SGD convergence by emphasizing harder examples towards the end of the training. This is useful for NLU since there are many frequent training examples that are \textit{easy} and can be deprioritize, e.g.\ ``play music'', ``how is the weather'', ``stop''. 

In Algorithm~\ref{sgd_ab}, we add active bias sampling to F10-SGD by replacing the random shuffle with sampling form discrete distribution $P_s(j)$ at specified epoch. $P_s(j)$ assigns higher probability to medium confidence examples, and lower probability to high and low confidence examples. We prioritize medium confidence examples and not low confidence ones, as the latter are more likely to be outliers.

\begin{algorithm}[H]

\caption{F10-SGD AB($F$, $\V{w}^0$, T)}
\label{sgd_ab}

\begin{algorithmic}[1]
\State atomic k $\gets$ 0
\For{e $\gets$ 0,1,..,Epochs or Convergence}
    
\If{$e <$ abStartEpoch}

    \State $S \gets $ shuffle(0..N)	
    
\Else
    \State $S \gets $ sample N inputs from $P_s$
    
\EndIf    
	\State ... The same as F10-SGD
\EndFor

\end{algorithmic}

\end{algorithm}
\vspace{-5pt}
We define $P_s(j)$ in Equation \ref{ab_prob}. $\bar{p}_h(\V{x}_j|\V{y}_j)$ is the average of the past $h$ iteration probabilities $p^k(\V{x}_j|\V{y}_j)$. The term $1 - \bar{p}_h(\V{x}_j|\V{y}_j)$ can be interpreted as the union of the alternative prediction probabilities. And, $\varepsilon$ is a smoothing prior that makes sure we assign non-zero probability to training examples with confidence close to one, otherwise the model may ``forget'' to recognize them.
\setlength{\abovedisplayskip}{3pt}
\setlength{\belowdisplayskip}{3pt}
\setlength{\abovedisplayshortskip}{0pt}
\setlength{\belowdisplayshortskip}{0pt}
\begin{align}
\label{ab_prob}
P_s(j) &= \bar{p}_h(\V{x}_j|\V{y}_j)(1 - \bar{p}_h(\V{x}_j|\V{y}_j)) + \varepsilon  \\
\varepsilon &= \ffrac{1}{N} \sum_j^N \bar{p}_h(\V{x}_j|\V{y}_j) (1 - \bar{p}_h(\V{x}_j|\V{y}_j)) \nonumber
\end{align}





In addition to sampling from $P_s$, we change the learning rate of the examples that we sampled. This way, we unbias the gradient estimation and ensure convergence \cite{zhao2015stochastic}. For each sample $j$, we have an importance weight $v_j = P_s(j) / V$ that is multiplied to the learning rate $\alpha$. Where $V = \ffrac{1}{N} \sum_j^N \bar{p}_h(\V{x}_j|\V{y}_j) (1 - \bar{p}_h(\V{x}_j|\V{y}_j)) + \varepsilon$ is normalization factor that makes $\ffrac{1}{N} \sum_j^N v_j = 1$, i.e.\ we do not change the global learning rate.

\section{Results}
\label{sec:experiments}

We evaluated F10-SGD on internal and public datasets. We used a Linux machine with Intel Xeon E5-2670 2.60 GHz 32 cores processor and 244 GB memory. The code is compiled with GCC 4.9 and -O3 optimization flags.

\textbf{Evaluation metrics.} We use accuracy for MaxEnt and F1 score for NER\footnote{\url{github.com/sighsmile/conlleval}}. We test for statistically significant differences using the Wilcoxon test \cite{hollander2013nonparametric} with 1000 bootstrap resamples and p-value < 0.05.

\subsection{Internal Datasets}

We evaluated F10-SGD against OWL-QN for DC, and the NER domains Music, Shopping and Cinema. The datasets details are in Table~\ref{datasets}. We use the development dataset for tuning the learning rate $\alpha$ and the L1/L2 hyper-parameters.

The reported speed improvements are only caused by the change in the optimizer. Since, the feature extraction, decoding, and model serialization are unchanged.


 \begin{table}[!hbtp]
{\footnotesize
\centering
\begin{tabular}{ l | c | c | c | c  }
\hline
Dataset & \#Labels & \#Train & \#Test & \#Dev \TBstrut\\
\specialrule{1.0pt}{0.5pt}{0.5pt}
DC & 24 & 50M & 600K & 600K \TBstrut\\
\hline 
Music NER & 67 & 6M & 140K & 140K \TBstrut\\

Shopping NER & 26 & 3.4M & 16K & 16K \TBstrut\\

Cinema NER & 27 & 3M & 3K & 3K \TBstrut\\
\hline
\end{tabular}
\caption{\label{datasets} Experimental datasets for the evaluation. DC uses 1 to 3 $n$-grams features. NER uses current word, left/right 1 to 3 grams, and gazetteers features. Gazetteers are lists of entities, such as artist names and song names.}
}
\end{table}
\vspace{-8pt}

\textbf{MaxEnt DC.} Table~\ref{maxent-dc} shows the MaxEnt DC results. The accuracy difference between F10-SGD and OWL-QN is not statistically significant, although F10-SGD accuracy is slightly higher. For DC, F10-SGD  AB does not bring additional accuracy improvements. The reason is that the model learns to assign high confidence to most of the sentences in the first 7 epochs of the training, and there are not many medium confidence sentences for the active bias sampling to prioritize. The training speeds of F10-SGD and F10-SGD AB are approximately 4x faster with half the number of CPU threads compared to OWL-QN. It takes only 10 epochs for SGD to reach the same accuracy level of OWL-QN with 250 epochs. Also, we get an additional 20\% speed up if we use 28 CPU threads.
\vspace{-5pt}
\begin{table}[!htb]
{\footnotesize
\centering
\setlength{\tabcolsep}{3pt}
\begin{tabular}{ l | c | c | c | c | c   }
\hline
Optimizer & CPU & Epoch & Size & Time & \%Acc \TBstrut\\
\specialrule{1.0pt}{0.5pt}{0.5pt}
OWL-QN & 28 & 250 & 202MB & 120m & 0 \TBstrut\\

F10-SGD & 14 & 10 & 192MB & 30m & 0.05 \TBstrut\\ 

F10-SGD & 28 & 10 & 192MB & 24m & 0.04 \TBstrut\\ 

F10-SGD AB & 14 & 7 + 3 & 200MB & 31m & 0.01 \TBstrut\\ 
\hline
\end{tabular}
\caption{DC results for OWL-QN vs F10-SGD and F10-SGD AB. Size is in megabytes. Time is the mean of 5 training runs in minutes. \%Acc is percentage accuracy relative change compared to OWL-QN (we are unable to disclose absolute accuracy). For F10-SGD AB we performed 7 normal epochs and 3 active bias epochs.}
\label{maxent-dc}
}
\end{table}

\vspace{-13pt}
Note that the Hogwild SGD does not scale linearly with the number of CPU threads~\cite{zhang2016hogwild++}. Training with 28 CPU threads is only 20\% faster than with 14 CPU threads. That is because of memory contention between the different threads for updating the shared parameter vector. When using more CPU threads, we get increased accuracy fluctuations from 0.02 to 0.04 across training runs.

\begin{table}[!hbtp]
{\footnotesize
\centering
\setlength{\tabcolsep}{3pt}
\begin{tabular}{ l | c | c | c | c | c  }
\hline
Optimizer &  CPU & Epoch & Size & Time & \%F1 \TBstrut\\
\specialrule{1.0pt}{0.5pt}{0.5pt}
\multicolumn{6}{c}{Music} \TBstrut\\
\hline

OWL-QN & 28 & 200 & 16MB & 164m & 0 \TBstrut\\

F10-SGD & 14 & 10 & 17MB & 41m & \hspace{1px} 0.02 \TBstrut\\ 

F10-SGD & 28 & 10 & 17MB &  26m & \hspace{1px} 0.02 \TBstrut\\ 

F10-SGD AB & 14 & 7 + 3 & 17MB & 41m & \hspace{1px} \textbf{0.42} \TBstrut\\ 
\hline
\multicolumn{6}{c}{Shopping} \TBstrut\\
\hline


OWL-QN & 4 & 200 & 16MB & 20m & \hspace{1px} 0  \TBstrut\\

F10-SGD    & 2 & 10 & 18MB & 5m & -0.24 \TBstrut\\ 

F10-SGD AB & 2 & 7 + 3 & 17MB & 5m &  \hspace{1px} 0.10 \TBstrut\\ 
\hline
\multicolumn{6}{c}{Cinema} \TBstrut\\
\hline

OWL-QN & 4 & 200 & 23MB & 19m & 0  \TBstrut\\

F10-SGD    & 2 & 10 & 22MB & 5m & \hspace{1px} 0.10 \TBstrut\\ 

F10-SGD AB & 2 & 7 + 3 & 22MB & 5m & \hspace{1px} \textbf{1.06} \TBstrut\\ 
\hline

\end{tabular}

\caption{\label{ner-crf-results} NER results for OWL-QN vs F10-SGD and F10-SGD AB. Size is in megabytes. Time is the mean of 5 training runs in minutes. \%F1 is percentage F1 relative change compared to OWL-QN (we are unable to disclose absolute F1 score). Bold numbers are statistically significant. For F10-SGD AB we performed 7 normal epochs and 3 active bias epochs.}
}
\end{table}
\vspace{-5pt}
\textbf{CRF Domain NER.} In Table~\ref{ner-crf-results}, the F1 score difference between OWL-QN and F10-SGD is not statistically significant for Music NER and Cinema NER, and statistically significantly worse for Shopping NER. F10-SGD AB statistically significantly improves F1 score for the Music and Cinema NER models, and for Shopping the F10-SGD AB difference with OWL-QN is not statistically significant. Similar to DC, the training speed of F10-SGD and F10-SGD AB is about 4x faster with half the number of CPU threads compared to OWL-QN. It takes only 10 epochs for F10-SGD AB to reach a comparable or better level of F1 score as 200 epochs with OWL-QN. Also, for Music NER, we get an additional 36\% speed up if we use 28 CPU threads. When training with 28 CPU threads, the F1 score fluctuation increases from 0.02 to 0.05.

We performed additional experiments on 24 domains, but for lack of space we do not present the results. The average training time reduction was 4x and the relative F1 score improvement of 0.5\%.



\subsection{Public Datasets}

We tested F10-SGD on public datasets against FastText (version from November 2018) for text classification and against CRFSuite (version 0.12) for NER. Both packages provide on of the fastest public implementations for their respective tasks.

\textbf{MaxEnt.} For the text classification evaluations, we used 8 datasets prepared by \citet{zhang2015character}. We tuned the hyper-parameters for MaxEnt and FastText on a part of the training data, and used 5 epochs and 4 threads for training. For FastText we used 10 word embedding dimension, which is the same as \citet{joulin2016bag}.

Table~\ref{maxent-vs-fasttext} shows the results. F10-SGD MaxEnt achieves 22\% relative speed-up compared to FastText, and it's faster to train on 7 out of 8 datasets. The MaxEnt models are faster to train on datasets with smaller number of classes, and the FastText models are comparable or faster with large number of classes because the dense output layer of FastText is CPU cache efficient. Furthermore, the MaxEnt models have smaller size compared to the FastText models because the former's vocabulary is pruned by the L1 regularization. The accuracy is comparable at 0.28\% average relative difference.

\begin{table}[!htb]
{\footnotesize
\setlength{\tabcolsep}{3pt} 
\centering
\begin{tabular}{ l | c c c | c c c   }
\hline 
 & \multicolumn{3}{c|}{FastText} & \multicolumn{3}{c}{F10-SGD} \TBstrut\\
 & Time & Size & Acc & Time & Size & Acc\TBstrut\\
\specialrule{1.0pt}{0.5pt}{0.5pt}
AG & 7s & 387MB & 92.44 & \bf3s & 11MB & \underline{92.46} \TBstrut\\
Sogou & 93s & 402MB & \underline{96.82} & \bf69s & 31MB & 96.49 \TBstrut\\
DBP & 26s & 427MB & 98.61 & \bf29s & 17MB & \underline{98.62} \TBstrut\\
Yelp P. & 44s & 408MB & 95.62 & \bf21s & 27MB & \underline{95.69} \TBstrut\\
Yelp F. & 51s & 411MB & \underline{63.94} & \bf46s & 147MB & 63.12 \TBstrut\\
Yah. A. & \bf95s & 494MB & \underline{72.44} & 100s & 275MB & 71.83 \TBstrut\\
Amzn. F. & 142s & 461MB & 60.33 & \bf127s & 300MB & \underline{60.35} \TBstrut\\
Amzn. P. & 164s & 471MB & 94.59 & \bf82s & 75MB & \underline{94.66} \TBstrut\\
\hline
\end{tabular}
\caption{DC results for FastText vs F10-SGD. Size is in megabytes. Time is the mean of 5 training runs in seconds. Both models use words and bigram features. Bold indicates fastest training, and underline indicates highest accuracy.}
\label{maxent-vs-fasttext}
}
\end{table}

\textbf{CRF.} For the NER evaluations, we used the CoNLL-2003 English NER dataset \cite{tjong2003introduction}, and the Ontonotes 5.0 English NER dataset \cite{weischedel2013ontonotes}. We tuned the CRF hyper-parameters on the dedicated development sets, and we used one thread for training because CRFSuite does not support training with multiple threads.

Table~\ref{ner-public-results} shows the results. F10-SGD CRF achieves 3.4x-3.7x relative speed-up compared to CRFSuite with the OWL-QN optimizer. We trained with 10 and 20 epochs for F10-SGD and 50 and 100 epochs for CRFSuite. For both algorithms less epochs degrade F1 and more epochs do not improve F1. The F1 score differences between the best configurations for F10-SGD and CRFSuite are not statistically significant. F10-SGD AB does not improve the F1 score on the public datasets because there are less ``easy'' examples to deprioritize compared to the internal NLU datasets.

\begin{table}[!hbtp]
{\footnotesize
\centering
\setlength{\tabcolsep}{3pt}
\begin{tabular}{ l | c | c | c | c  }
\hline
Optimizer & Epoch & Size & Time & F1 \TBstrut\\
\specialrule{1.0pt}{0.5pt}{0.5pt}
\multicolumn{5}{c}{CoNLL 2003 \#Labels:\ 4 \#Train:\ 14986} \TBstrut\\
\hline
CRFSuite & 50 & 12MB & 2.8m & 83.39 \TBstrut\\
CRFSuite & 100 & 9.5MB & 4.9m & 83.34 \TBstrut\\
F10-SGD & 10 & 9.7MB & 0.75m & 83.35 \TBstrut\\ 
F10-SGD AB & 7 + 3 & 9.6MB & 0.8m & 83.46 \TBstrut\\
F10-SGD & 20 & 8.8MB & 1.4m & 83.34 \TBstrut\\
\hline
\multicolumn{5}{c}{Ontonotes 5.0 \#Labels:\ 18 \#Train:\ 107973} \TBstrut\\
\hline


CRFSuite & 50 & 68MB & 102m & 79.49 \TBstrut\\
CRFSuite & 100 & 49MB & 187m & 79.99 \TBstrut\\
F10-SGD & 10 & 50MB & 30m & 79.50 \TBstrut\\ 
F10-SGD AB & 7 + 3 & 52MB & 31m & 79.98 \TBstrut\\
F10-SGD & 20 & 40MB & 52m & 79.60 \TBstrut\\

\hline

\end{tabular}

\caption{\label{ner-public-results} NER results for OWL-QN vs F10-SGD and F10-SGD AB. Size is in megabytes. Time is the mean of 5 training runs in minutes. The features used are current word, left/right 1 to 3 grams.}
}
\end{table}

\vspace{-13pt}


\section{Conclusions}
\label{sec:Conclusion}
Fast model training is important for continuous improvement of voice-assistant's NLU. However, training times are increasing as our datasets are growing. To reduce model training time, we developed the F10-SGD optimizer that can train elastic-net linear models for text classification and NER up to 4x faster compared to OWL-QN. This is accomplished without loss of accuracy or increase in model size. In addition, F10-SGD with active bias sampling provides small but statistically significant improvements in NER F1 scores on NLU datasets.



\bibliography{fast_train}
\bibliographystyle{acl_natbib}

\end{document}